\pdfoutput=1
\documentclass[11pt]{article}

\usepackage[final]{acl}

\usepackage{times}
\usepackage{latexsym}

\usepackage[T1]{fontenc}
\usepackage[utf8]{inputenc}

\usepackage{microtype}

\usepackage{inconsolata}

\usepackage{graphicx}

\usepackage{soul,color}
\usepackage{booktabs}
\usepackage{verbatim}
\usepackage{multicol}
\usepackage{listings}
\usepackage{amsmath}
\title{UrbanLLM: Autonomous Urban Activity Planning and Management with Large Language Models}

\author{
 \textbf{Yue Jiang\textsuperscript{1,3}},
 \textbf{Qin Chao\textsuperscript{1,3}},
 \textbf{Yile Chen\textsuperscript{1*}},
 \textbf{Xiucheng Li\textsuperscript{2}},
 \textbf{Shuai Liu\textsuperscript{1}},
 \textbf{Gao Cong\textsuperscript{1*}},\\
 \textsuperscript{1}Nanyang Technological University, Singapore,\\
 \textsuperscript{2}Harbin Institute of Technology(Shenzhen), China,\\
 \textsuperscript{3}DAMO Academy, Alibaba group, Singapore,
\\
 \small
   \textbf{\{yue013@e, chao0009@e, yile001@e, shuai004@e, gaocong@\}ntu.edu.sg}\\
   \small\textbf{lixiucheng@hit.edu.cn}
 }

\begin{document}
\maketitle
\begin{abstract}
Location-based services play an critical role in improving the quality of  our daily lives. Despite the proliferation of numerous specialized AI models within spatio-temporal context of location-based services, these models struggle to autonomously tackle problems regarding complex urban planing and management. To bridge this gap, we introduce \textbf{UrbanLLM}, a fine-tuned large language model (LLM) designed to tackle diverse problems in urban scenarios. UrbanLLM functions as a problem-solver by decomposing urban-related queries into manageable sub-tasks, identifying suitable spatio-temporal AI models for each sub-task, and generating comprehensive responses to the given queries.
% Urban activity planning and management is important in our daily live. Despite the availability of numerous spatio-temporal AI models, these models struggle to autonomously handle intricate urban tasks, demanding significant effort from experienced researchers and software engineers. By leveraging the capabilities of large language models (LLMs) in language comprehension, and reasoning, we introduce \textbf{UrbanLLM}, a fine-tuned LLM for autonomous urban activity planning and management. UrbanLLM is designed to decompose real-world urban activity planning tasks into manageable sub-tasks and identify suitable spatio-temporal machine learning models for each task, thereby enhancing the accuracy of urban planning and the efficiency of management processes. UrbanLLM operates through three stages: spatio-temporal task analysis, model matching, and results generation.
% Each stage is supported by extensive engineering design to ensure smooth automation and integration. 
Our experimental results indicate that UrbanLLM significantly outperforms other established LLMs, such as Llama and the GPT series, in handling problems concerning complex urban activity planning and management.  UrbanLLM exhibits considerable potential in enhancing the effectiveness of solving problems in urban scenarios, reducing the workload and reliance for human experts. Our code is available at: https://anonymous.4open.science/r/UrbanLLM-1227/
% (Due to the double-blind reviewing process, code will be released after review phase)
\end{abstract}
% \vspace{-0.5cm}
\section{Introduction}
% \vspace{-0.2cm}
Location-based services are ubiquitous in urban spaces, supporting a range of scenarios from commuting assistance for travelers and daily activities for residents to event monitoring for city regulators. The diverse and substantial demand for location-based services has driven the development of specialized AI models tailored to specific tasks within spatio-temporal context. These tasks include (spatio-)temporal forecasting \cite{AGCRN,MTGNN,DCRNN}, imputation \cite{BRITS,liushuai}, and anomaly detection \cite{detectioniclr, detectionicml}, as well as travel time estimation \cite{ETACIKM, ETAWWW}, trajectory prediction \cite{MtrajREC,RNtrajREC}, and POI recommendation \cite{POIrecICDE,POIrecSIGIR}, etc. 

Despite the promising results of specialized AI models in addressing urban-related tasks with fixed input formats, these models are inadequate for handling complex queries in natural language that require strategic planning, reasoning, and collaboration among multiple specialized models across potentially various data modalities, such as GPS coordinates, addresses, and traffic records. As illustrated in Figure 1, to answer the given query, human experts need to decompose it into several sub-tasks, select and employ an appropriate combination of models for each sub-task, and synthesize the response based on model outputs. To enhance user experiences in location-based services, it is critical to design an autonomous and effective method for solving urban-related problems, particularly regarding complex urban activity planning and management.

\begin{figure}[t]
  \includegraphics[width=\columnwidth]{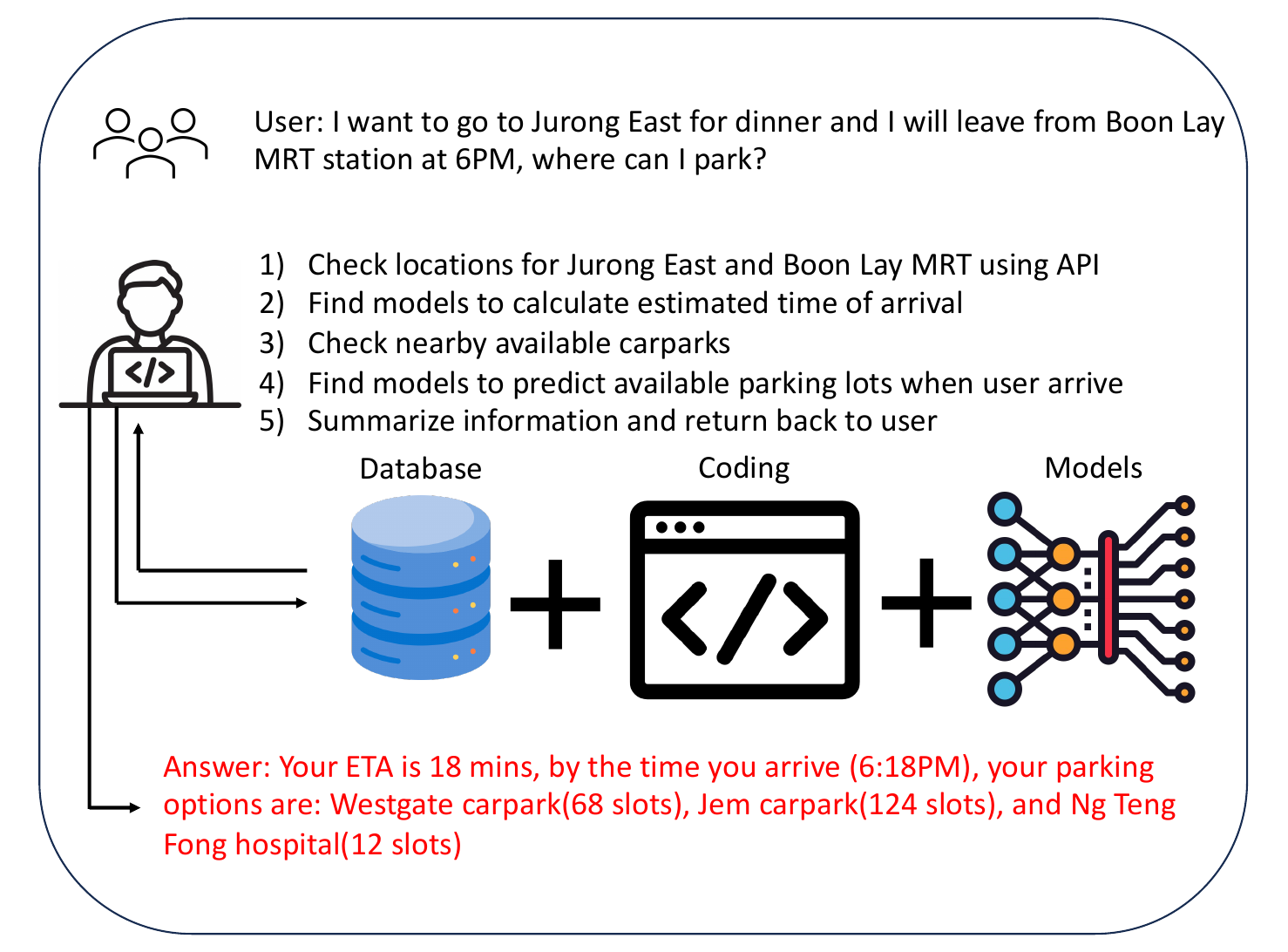}
  \caption{The process of solving a real-world problem in urban scenarios involving urban activity planning by human experts.}
  \label{fig:experiments}
\end{figure}

Large language models (LLMs), such as ChatGPT \footnote{https://platform.openai.com/docs/models}, have attracted significant interest from both academia and industry \cite{GeoLLM, hugginggpt}. LLMs have been applied in diverse domains, such as commerce, finance, healthcare, and the geospatial domain \cite{LLMdomain,LLMdomainNIPS,geogpt,K2}. Their exceptional capabilities in language comprehension and reasoning have positioned them as core modules in autonomous agents, where they function as problem solvers~\cite{hugginggpt}.
% AutoGPT~\cite{} and AgentGPT~\cite{}. 
However, conventional LLMs, as well as LLM agents built on them, encounter several challenges specific to our target.
\underline{First}, conventional LLMs, despite their intrinsic ability in reasoning and comprehension, possess limited geospatial knowledge about tasks and AI models within spatio-temporal context in urban scenarios~\cite{GeoLLM}.
\underline{Second}, agents such as AutoGPT \footnote{https://github.com/Significant-Gravitas/Auto-GPT}, AgentGPT \footnote{https://github.com/reworkd/AgentGPT}, and BabyAGI \footnote{https://github.com/yoheinakajima/babyagi} utilize tools for web search and code execution, which are not suited for location-based services. Moreover, they are designed to refine solutions for individual tasks in a recursive way, lacking the ability to holistically analyze and decompose urban-related problems for multiple specialized models. 
\underline{Third}, while HuggingGPT \cite{hugginggpt} proposes to schedule text and image-related problems into sub-tasks and coordinate responses from corresponding models, it still relies on conventional LLMs as its backbones, thus inheriting their limitations in solving tasks in urban scenarios.

To tackle these challenges, we propose UrbanLLM, a novel urban foundation model to effectively decompose queries into sub-tasks and schedule specialized AI models within spatio-temporal context for each sub-task, thereby autonomously solving complex problems in urban scenarios. The idea is to align the problem-solving process with the established paradigms of LLMs, and enhance LLMs with the capability for urban activity planning and management through targeted fine-tuning. In this process, LLMs are equipped to serve as a universal interface capable of handling diverse tasks for various urban scenarios and producing appropriate responses.
% To address these above-mentioned limitations, we propose that, after proper fine-tuning, LLMs can function as a supervisory entity to schedule existing machine learning methods to address complex urban activity planning tasks autonomously. In this context, language would serve as a universal interface to facilitate this process. This proposal aims to be the first to deliver UrbanLLM, an urban foundation model to break down complex practice urban activity planning tasks into smaller, manageable tasks and identify the relevant deep learning models for each task autonomously, thereby, empowering UrbanLLM to solve complex urban task from multiple domains and modalities.

Specifically, LLMs are initially fine-tuned using a structured template with high-quality examples that focus on spatio-temporal task decomposition and scheduling, thus augmenting the reasoning capabilities for the problems in targeted urban scenarios. Subsequently, the processing of new urban-related problems is conducted in three stages for inference phase. First, in the task analysis stage, a query of urban-related problems is effectively decomposed into a series of sub-tasks, each of which corresponds to a specific type of specialized AI models within spatio-temporal context. Next, in the model matching stage, the most suitable model is selected for each sub-task from a pool of candidate specialized AI models based on their descriptions. Finally, in the results generation stage, the selected models are executed to obtain output results, which are then organized into prompts to formulate a comprehensive response to the original query. By utilizing UrbanGPT, we facilitate the efficient and convenient resolution of urban-related problems without intensive manual efforts. Our contributions are summarized as follows.

% To be specific, UrbanLLM is first fine-tuned in the learning phase to adopt spatial-temporal task decomposition ability and contains three stages in the inference phase: a spatio-temporal task analysis stage to decompose user request into a series of spatial-temporal tasks, a model matching stage to pair each spatial-temporal task with potential spatial-temporal deep learning methods and select the most appropriate one, and a results generation stage to run the selected spatial-temporal methods and formulate results into a prompt and return to users. The objective of our work is to enhance efficiency and effectiveness in urban planning and management with the help of LLMs. As the community continues to develop an increasing number of dedicated machine learning models, UrbanLLM will enhance its adaptability in urban-related planning and management tasks. It is poised to become the comprehensive, go-to machine learning model for a multitude of urban tasks without the need for experienced researchers/software engineers. In summary, our contributions are as follows.
\begin{itemize}
% \vspace{-0.1mm}
% \item Despite the challenges associated with crafting prompts, which are often sensitive to small changes \cite{promptcraft}, we effectively developed prompts for instruction-tuning the UrbanLLM, and the spatial-temporal inference phase to employ UrbanLLM as a universe solver.
\item We propose UrbanLLM, the pioneering application of a fine-tuned LLM based on Llama-2-7B \cite{llama2}, to solve problems regarding complex urban activity planning and management. UrbanLLM is expected to improve in performance and adaptability as the community continues to developing AI models within spatial-temporal context.
\item We develop an effective method for LLM instruction-tuning that enhances the reasoning capabilities in urban scenarios. Furthermore, we devise an autonomous pipeline to generate responses by minimizing the need for human intervention.    
\item Extensive experiments on real-world problems in urban scenarios demonstrate that our proposed UrbanLLM significantly outperforms other advanced LLMs, such as Llama-3-8B \footnote{https://huggingface.co/meta-llama/Meta-Llama-3-8B} and GPT-4o \footnote{https://platform.openai.com/docs/models/gpt-4o}, by a substantial margin.
% \item We first introduce UrbanLLM, a fine-tuned LLM based on Llama-2-7B, designed to address real-world urban activity planning and management problems without requiring extensive resources from experienced researchers and software engineers. Leveraging our meticulously crafted instruction-tuning dataset and comprehensive spatial-temporal model zoo, UrbanLLM is poised to enhance its performance and adaptability in a broader range of urban-related planning and management tasks as the community continues to develop additional spatial-temporal machine learning models.

% \vspace{-0.2mm}
\end{itemize}

% \textcolor{blue}{The primary challenges in this project include: 1. Large language models are currently unable to extract and comprehend geospatial semantics, a fundamental requirement for effectively utilizing these models in downstream urban planning tasks. 2. A significant challenge lies in the fact that while large language models can analyze task decomposition and planning akin to human cognition, the vast scope and diverse types of urban-related tasks may lead to issues. Specifically, querying these models for task decomposition and planning could potentially result in the 'LLM hallucination' problem, thereby providing incorrect solutions. 3. Despite large language models' ability to analyze task decomposition and planning akin to humans, they still require demonstrations for causal analysis. Consequently, the involvement of human experts becomes necessary to provide accurate examples of task analysis, task planning, and machine learning model selection to the foundation model. However, this process is resource-intensive.}
% \section{Related Works}
% \vspace{-0.5cm}
\section{Related Work}
Large language models (LLMs), due to their powerful capability in reasoning and comprehension, are increasingly utilized to assist in specific urban applications. For example, TrafficGPT \cite{trafficGPT} employs ChatGPT as a control agent to interact with various system components, such as databases, visualization and statistical tools, to perform basic analytical operations in traffic-related tasks. GeoGPT \cite{geogpt} utilizes ChatGPT to address similar analytical operations in Geographical Information Systems (GIS) domain. \cite{Purbanplanning} leverages LLMs to simulate roles such as planners and residents within a multi-agent framework to help urban land use and development planning. LLMob\cite{LLMob} introduces a framework that considers individual activity patterns for urban mobility data generation. However, these studies typically rely on original LLMs and possess limited domain knowledge, which restricts their effectiveness in addressing challenges regarding complex urban planning and management. 

% TrafficGPT \cite{trafficGPT} employs ChatGPT as a control agent to interact with models that tackle traffic tasks, which consist of traditional statistical methods and tools, to handle urban traffic-related tasks recursively. GeoGPT \cite{geogpt} directly uses the GPT-3.5 model as an autonomous control agent for data collection, analysis, and visualization tasks in Geographical Information Systems (GIS). \cite{Purbanplanning} proposes utilizing LLMs in roles such as planners and residents within a multi-agent framework to assist in urban land use and development planning. Similarly, \cite{LLMob} introduces an LLM agent framework that considers individual activity patterns and motivations for urban mobility data generation.
To overcome such intrinsic limitations, various initiatives have deployed fine-tuned LLMs in targeted scenarios. For example, LLMlight \cite{LLMlight} is fine-tuned based on Llama-2 to improve decision-making and policies in traffic signal control. TransGPT \cite{transGPT} is trained on a corpus of examples from traffic prediction, public transportation, and autonomous driving, to address urban transportation tasks such as identifying traffic rules and signs. UrbanGPT \cite{UrbanGPT} aims to enhance spatio-temporal forecasting accuracy in a zero-shot setting by integrating a specialized decoder with an instruction-tuning paradigm. GeoLLM \cite{GeoLLM} enhances urban regional questions such as population density and home value by fine-tuning LLMs with specifically designed templates. 
Moreover, several models focus more on text-related dimensions. PlanGPT \cite{PlanGPT} is fine-tuned on a large corpus of urban planning regulations from numerous local governments in China, aimed at revising or generating texts for new regulations and evaluating planning documents.
K2 \cite{K2} learns additional geospatial knowledge from a collection of geoscience text training corpus, enhancing NLP tasks such as summarization and text classification, specifically for the geoscience domain. 
Different from all these studies, we are the first to equip LLMs with the capability to decompose queries regarding complex urban planning and management into manageable components that align with specialized AI models within spatio-temporal context, thereby autonomously tackling diverse urban-related problems.

% Currently, no fine-tuned LLMs incorporate geospatial knowledge and state-of-the-art dedicated spatio-temporal AI models as backend executors to address real-world urban planning and management scenarios autonomously. To bridge this gap, we introduce UrbanLLM, a model specifically designed to tackle these challenges.
% \vspace{-0.5cm}
\section{UrbanLLM}
% \vspace{-0.5cm}
% We believe that an autonomous urban activity planner and manager should possess the following capabilities: First, it should be able to understand the complex transportation problems raised by urban users and decompose them into interdependent spatial-temporal tasks. Secondly, it should be able to match suitable models from an existing spatio-temporal models zoo for these subtasks and provide the appropriate inputs for the different models. Finally, it should be able to correctly combine the results of these models to present the final solution. 

% Based on those requirements, we have designed UrbanLLM, which consists of two phases: the learning phase and the inference phase. 

Our proposed UrbanLLM is structured into two phases: the learning phase and the inference phase. In the learning phase, we fine-tune UrbanLLM, employing meticulously crafted examples from our constructed self-instruct dataset. This dataset includes examples that contain reasoning hints, various types of backbone spatio-temporal AI models, diverse queries, and the corresponding decomposed sub-tasks. This fine-tuning process effectively enhances the comprehension and reasoning capabilities of UrbanLLM tailored to address our targeted objectives in urban scenarios. 

\begin{figure*}[h]
  \includegraphics[width=\textwidth]{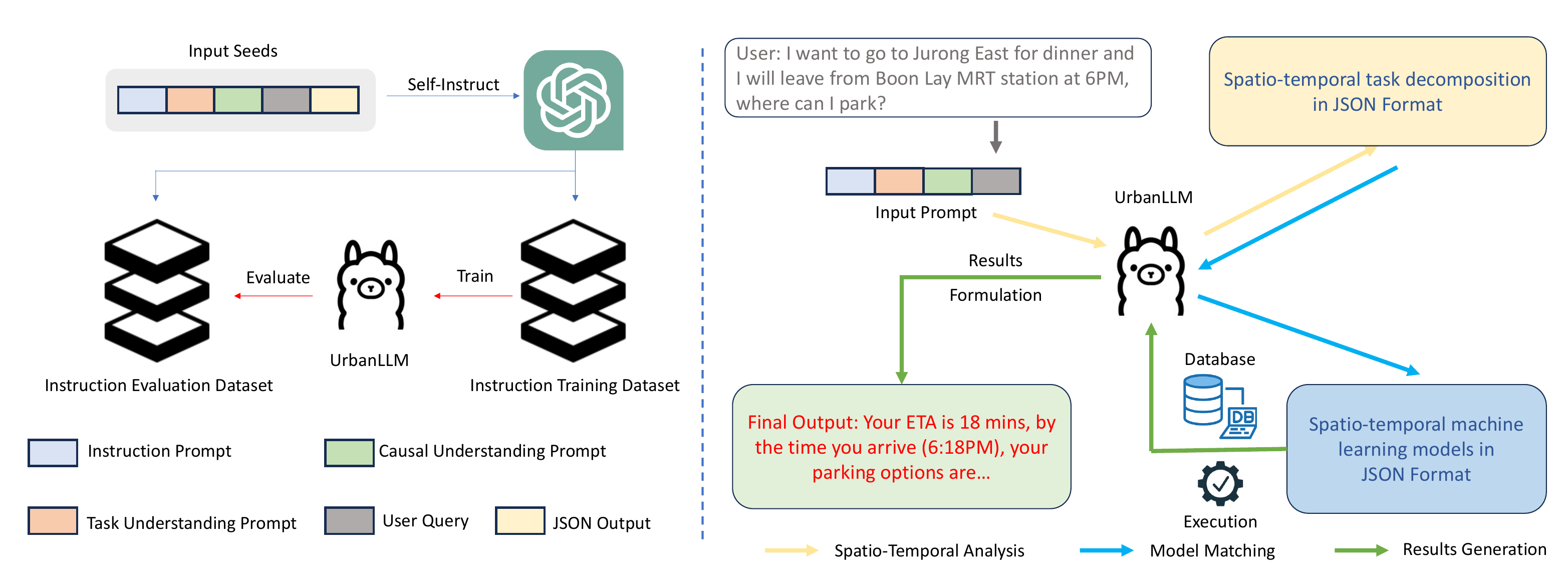}
  \caption{The overall process of the proposed UrbanLLM framework. The urban activity planning learning phase is on the left and the inference phase is on the right.}
  \label{fig:framework}
\end{figure*}

The inference phase consists of three stages: spatial-temporal analysis, model matching, and results generation. In the spatio-temporal analysis stage, UrbanLLM receives queries that are organized within the same prompt template used during the training stage with new queries, enabling it to effectively decompose the query into a series of types of spatio-temporal tasks, owing to enhancements achieved through fine-tuning. The model matching stage involves pairing each identified spatio-temporal task with suitable AI models and selecting the most appropriate one. Finally, in the results generation stage, the selected spatio-temporal models are executed, and their outputs are formulated into a prompt to produce the response to the query. The overall process for the two phases is depcited in Figure \ref{fig:framework}.

% To be specific,  UrbanLLM starts from fine-tuning a Llama-2-7B model using a meticulously designed instruction prompt with a self-instruct dataset.
% This learning phase enhances the model's comprehension of spatial-temporal tasks, allowing it to decompose the real-world scenarios posed by users into subtasks.

% The inference phase consists of three stages: spatial-temporal analysis, model matching, and results generation. The spatio-temporal task analysis stage decomposes the user request into a series of spatio-temporal tasks. The model matching stage pairs each spatio-temporal task with potential spatio-temporal deep learning methods and selects the most appropriate one. Finally, the results generation stage executes the chosen spatio-temporal methods, formulates the results into a prompt, and returns them to the users. The overall framework is illustrated in Figure \ref{fig:framework}.
% We have carefully designed instruction prompts for different phases, and a detailed example is shown in Figure \ref{fig:prompts}.
% Overall, UrbanLLM provides a robust and efficient framework for tackling complex planning challenges.
% \vspace{-0.2cm}
\subsection{Urban Activity Planning Learning}\label{sec:learning}
% \vspace{-0.2cm}
% Our first objective is to equip LLMs with the capability to comprehend knowledge about different types of spatio-temporal tasks and decompose complex queries scenarios into these tasks. This is achieved by rigorously formulating the instructions and prompts and subsequently fine-tuning a Llama2-7B model (Figure \ref{fig:framework} left). 

The learning phase of UrbanLLM is designed to endow LLMs with the capability to comprehend knowledge for processing different types of spatio-temporal tasks, facilitating the decomposition of queries in urban scenarios into these tasks. This is achieved through the rigorous formulation of prompts and instructions, followed by the fine-tuning of a Llama2-7B model (Figure \ref{fig:framework} left). 

\begin{figure*}[h]
  \includegraphics[width=\textwidth]{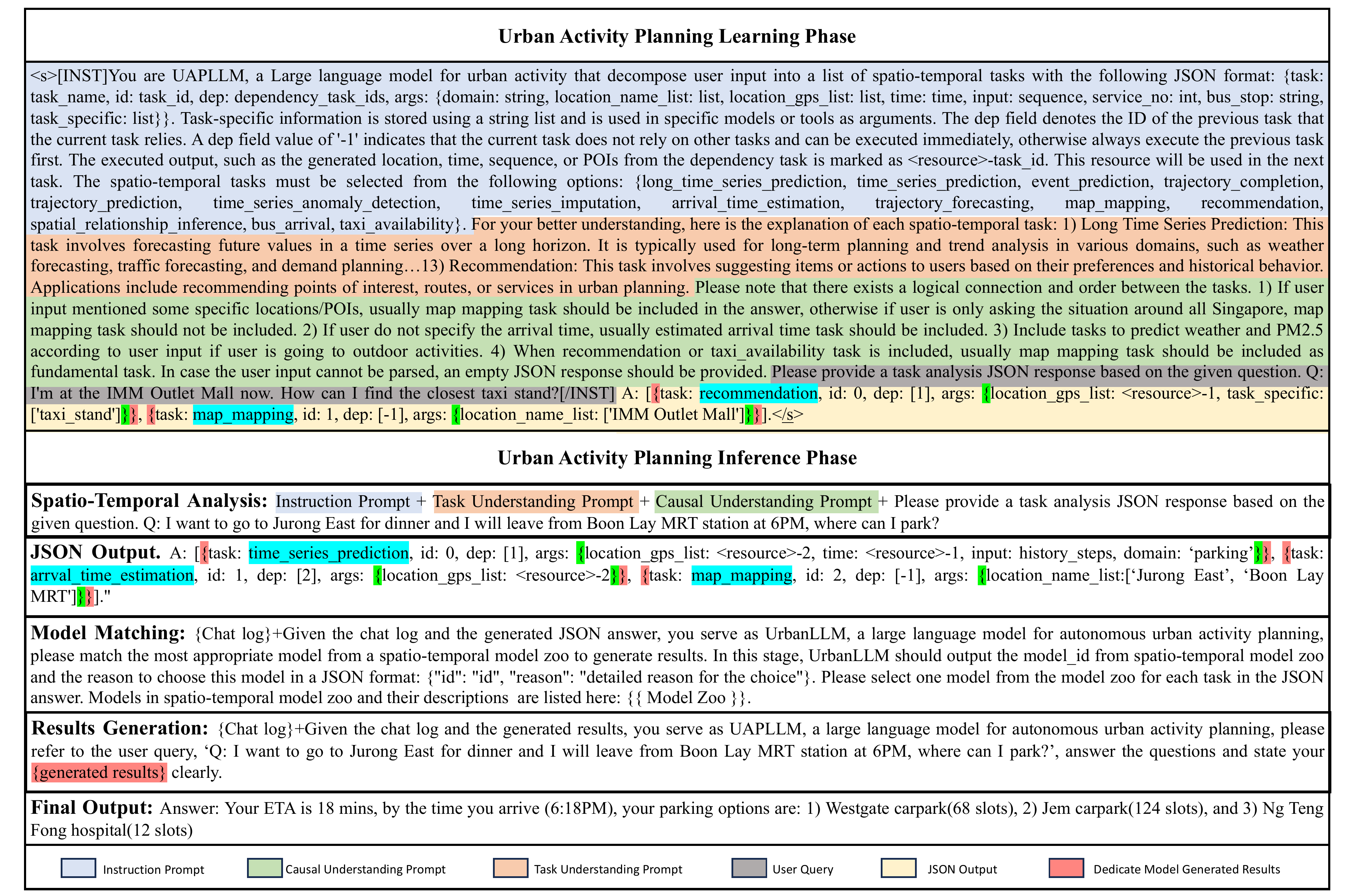}
  \caption{A sample of detailed prompt template used in the learning phase and the process of a new query solved by UrbanLLM in the inference phase.}
  \label{fig:prompts}
\end{figure*}

Specifically, we recorded 170 seed prompts in Singapore from human experts and employed the self-instruct method \cite{selfinstruct} to generate additional 15,249 training examples and 1,694 evaluation examples using \texttt{GPT-4-1106-preview}. The training examples are utilized for instruction-tuning UrbanLLM in an unsupervised learning paradigm, while the evaluation examples serve to evaluate the performance of the all compared models in the experiments. Each example in the dataset comprise an instruction part and a QA part, with an sample showcased in Figure~\ref{fig:prompts}. 

The instruction part features three key components: scenario formulation, task understanding, and causal understanding. Scenario formulation specifies the requirements of translating task decomposition into a machine-understandable format (i.e., JSON), and defines 13 types of potential spatio-temporal sub-tasks and their associated arguments in the format, resulting in a total of 34 task combinations (listed in Appendix \ref{sec:taskscombinations}). Inspired by HuggingGPT \cite{hugginggpt}, to demonstrate the dependency relationship among tasks, we use the \texttt{`dep'} field to denote the task ID of a previous task upon which the current task relies, and the \texttt{<resource>-task\_id} to indicate the output from the previous task used as the input for the current task. Task understanding provides detailed explanations on each type of spatio-temporal task (listed in Appendix \ref{sec:taskunderstanding}), enabling UrbanLLM to understand their functions for subsequent query decomposition. Causal understanding identifies the connections and causal relationships among specific task combinations, allowing UrbanLLM to grasp and apply the underlying logic. The QA part includes the specific queries regarding complex urban activity planning and management, and its response adhering to the JSON format specified in the instruction section.
Subsequently, we utilized the constructed training examples to fine-tune the Llama-2-7B model with QLoRA \cite{quanti, lora}, a technique that significantly reduces the computational cost of fine-tuning. Through the fine-tuning process, UrbanLLM learns to adeptly schedule urban activity planning and management for diverse and complex queries in urban scenarios.
% The pre-trained LLMs lack a sufficient training corpus related to urban activity planning and management scenarios, leading to the knowledge shortcut problem and causing the LLMs to exhibit faithfulness hallucination \cite{selfinstruct}.
% Detailed cases of this issue are provided in Appendix \ref{sec:appendix}. 

% To address this challenge, we instruction-tuned UrbanLLM based on the Llama-2-7B model using QLoRA \cite{quanti, lora}, a technique that significantly reduces the computational cost of fine-tuning. This tuning process utilized previously generated training samples, which included both instruction and QA sections, within an unsupervised learning framework.

% \vspace{-0.2cm}
\subsection{Spatio-Temporal Analysis}
% \vspace{-0.2cm}
In the first stage of the inference phase, UrbanLLM employs the template in the training phase to craft a prompt. This prompt is then fed into the fine-tuned model to produce JSON outputs which present the results of spatio-temporal task analysis. Specifically, these outputs provide structured information on the dependencies and interactions among the 13 defined sub-tasks, outlining the task decomposition necessary to address the given query. Through the implementation of fine-tuning in Section 3.1, UrbanLLM gains the knowledge to decompose the query in urban scenarios into manageable spatio-temporal sub-tasks, each associated with its specialized AI models. This decomposition is critical for solving complex problems regarding urban activity planning and management, which are typically exceed the capability of single models. Finally, the generated JSON outputs are utilized in the subsequent stages of model matching and results generation.
% \vspace{-0.2cm}
\subsection{Model Matching}
% \vspace{-0.2cm}
In the model matching stage, the chat log from previous interactions is used as input for UrbanLLM to select the appropriate model for each sub-task. To facilitate this process, we have organized a comprehensive model zoo consisting of more than 50 recent spatio-temporal AI models and tools (some are presented in Appendix \ref{sec:modelmatching}). Each model is associated with descriptions that cover model information on the addressed problem settings and scenarios, as well as the data types or formats in which the model has been tested or evaluated. Utilizing these descriptions, UrbanLLM is prompted to match each sub-task identified during the spatio-temporal analysis stage with a suitable model from the model zoo. Finally, UrbanLLM outputs the selection of the most suitable spatio-temporal model for each sub-task in JSON format, ensuring accurate and efficient task execution.
% In the model matching stage, the spatio-temporal analysis results generated in the previous phase are input into UrbanLLM as a chat log. To support this process, we have compiled descriptions of over 50 recent spatio-temporal machine learning methods and basic tools, which are stored in our spatio-temporal model zoo. Detailed descriptions of these spatio-temporal machine learning methods are presented in Appendix \ref{sec:modelmatching}. We assert that while words may have consistent meanings across different domains (word vocabulary \cite{word2vec}), time series data and trajectory data from various domains exhibit unique natural dynamics and cannot share a universally effective embedding. Consequently, we specifically annotate the data domain in which each spatio-temporal machine learning method has been tested or evaluated. During the model matching stage, UrbanLLM selects appropriate methods from the spatio-temporal model zoo based on the task types identified in the spatio-temporal analysis stage. These candidate methods are then prompted into UrbanLLM, which matches the spatio-temporal descriptions of the candidate models to the task instructions from the chat log. This process outputs the most suitable spatio-temporal method for each task in JSON format, ensuring accurate and efficient task execution.
% \vspace{-0.5cm}
\subsection{Results Generation}
% \vspace{-0.2cm}
In the results generation stage, UrbanLLM executes the selected spatio-temporal models determined by the JSON output from the model matching stage. The execution sequence follows the dependencies outlined in the JSON output from the spatio-temporal analysis stage. For each model, UrbanLLM retrieves necessary inputs either directly from the specified arguments,  or from the outputs of previously executed models where dependencies exist. When the last model in the execution sequence is accomplished, the results are aggregated and compiled into a prompt that is then processed by UrbanLLM to generate the response. In this way, we ensure that the spatio-temporal models are logically executed, producing a final response to the initial query.
% In the Results Generation stage, UrbanLLM executes the spatio-temporal methods based on the JSON result from the model matching stage. For each specific spatio-temporal method, UrbanLLM retrieves the necessary input data from the database by identifying arguments containing spatial and temporal keywords from the spatio-temporal analysis output JSON. This input data is used to generate intermediate results, which are subsequently used as input for the next spatio-temporal method in the sequence. If there is no subsequent spatio-temporal method, the final results are compiled into a prompt format and returned to the user by UrbanLLM. This systematic approach ensures that the spatio-temporal methods are executed in a logical and efficient sequence, producing accurate and actionable results for urban activity planning management.
% \vspace{-0.4cm}
\section{Experiments and Results Discussion}
% \vspace{-0.2cm}
In this section, we compare UrbanLLM with several strong LLM baselines, such as Llama-3 and GPT-4o, to demonstrate the urban activity planning and management capability and superior performance of UrbanLLM after instruction fine-tuning. We then include a case study to visually illustrate the differences between UrbanLLM and GPT-4o. Furthermore, we conduct an  ablation study to test the contribution of each component in UrbanLLM.
% \vspace{-0.2cm}
\subsection{Experimental Setup}
% \vspace{-0.2cm}
\noindent\textbf{Datasets}. We developed our UrbanLLM based in Singapore where extensive urban data sources are available. We recorded 170 seed prompts and applied the self-instruct method to generate 15294 training examples and 1694 evaluation examples using \texttt{GPT-4-1106-preview}. The training examples contains 4787 simple queries, which require a single model to derive the results, and 10507 complex requests, which require coordinating among multiple models for problem-solving. The evaluation examples contain 427 simple queries and 1267 complex queries. In addition to the self-instruct evaluation dataset, we have further constructed a human-annotated dataset consisting of 200 queries with corresponding responses focused on scenarios concerning complex urban activity planning and management, to rigorously evaluate UrbanLLM's performance. The data sources which serve as inputs for the specialized spatio-temporal AI models employed in the results generation stage are retrieved from the Singapore Open Data API, which provides access to over 4000 datasets from 69 government agencies. This API offers diverse spatio-temporal data sources from various domains, such as bus locations, POIs, passenger flow, car park availability, precipitation records, and PM2.5 levels.

\noindent\textbf{Baselines}. We apply our UrbanLLM, and several LLMs serving as baselines in the inference phase to evaluate their performances:

\begin{itemize}
    \item \textbf{Llama-2-7B}: Llama-2-7B is a open-source LLM developed by Meta AI with 7 billion parameters.
    \item \textbf{Vicuna-7B-v1.5\footnote{https://huggingface.co/lmsys/vicuna-7b-v1.5-16k}}: Vicuna-7B-v1.5 is a open-source LLM based on Llama-2-7B with additional fine-tuning and supporting 16k context length.
    \item \textbf{Llama-3-8B}: Llama-3-8B is the latest model in the Llama series from Meta AI, which features an expanded architecture with 8 billion parameters. This model offers further enhancements in processing power and language comprehension.
    \item \textbf{GPT-3.5\footnote{https://platform.openai.com/docs/models/gpt-3-5-turbo}}: GPT-3.5 model  is a chatbot-based LLM (\texttt{gpt-3.5-turbo-0613}) developed by OpenAI. As the model is unavailable, we execute the inference phase using its API.
    \item \textbf{GPT-4o}: GPT-4o is a more advanced iteration of the GPT series after GPT-3.5. Similarly, we we execute the inference phase using its API.
\end{itemize}

\noindent\textbf{Evaluation Metrics}. We employ four metrics: accuracy, precision, recall, and F1 score, to evaluate the performance for each evaluation example and report the weighted average result. Accuracy is calculated as the proportion of predicted examples (including sub-tasks and their dependencies) that exactly match the ground truth among the total number of  evaluated examples. Precision, recall, and F1 score are computed at a micro level for each example, specifically measuring sub-task predictions. More details of these evaluation metrics can be found in Appendix~\ref{sec:evluationmetrics}.
% also known as the positive predictive value, measures the accuracy of the positive predictions. Recall, or sensitivity, quantifies the ability of the model to identify all relevant samples. The F1 score is the harmonic mean of precision and recall.

\noindent\textbf{Implementations}. UrbanLLM is fine-tuned on the training examples for 5 epochs on a Linux workstation with an Intel(R) Xeon(R) Gold 6248 CPU @ 2.50GHz and 8 32GB Tesla V100 GPU. We used 4-bit quantization \cite{quanti} to obtain a more compact model representation, and low rank adaptation (LoRA) \cite{lora} to reduce the number of trainable parameters and decrease the GPU memory requirements. We set LoRA attention dimension to be 64 and initial learning rate to be 2e-4 with Adam optimizer.

% \vspace{-0.3cm}
\subsection{Performance of UrbanLLM}
% To evaluate the per of LLMs, we first calculate the accuracy (exact match), precision, recall, and F1 score for each evaluation sample and report the weighted average results (listed in Appendix \ref{sec:evluationmetrics}) across all evaluation dataset for all baselines and our UrbanLLM in Table \ref{tab:results}. Additionally, we present their performance on both simple and complex real-world scenarios in Table \ref{tab:simple} and Table \ref{tab:complex}.
% \vspace{-0.2cm}
To evaluate the analytical capabilities of LLMs, we report the metrics for all evaluated scenarios across all compared models in Table \ref{tab:results}. In addition, we present their detailed performance in both simple and complex real-world examples in Table \ref{tab:simple} and Table \ref{tab:complex}, respectively. The best result for each evaluation metric is highlighted in bold and the second best result is highlighted with an underline.

In these evaluation examples, we generally observe that baseline LLMs show excellent performance in completing both the model matching and results generation stages, echoing the observations from previous research\cite{hugginggpt}. However, a notable deficiency arises in the spatio-temporal analysis stage for the baseline LLMs due to the limited urban-specific training corpus.  
During this critical stage, baseline LLMs frequently encounter hallucination issues, such as generating non-existent tasks. Since this stage is crucial in translating queries into corresponding spatio-temporal sub-tasks to be solved using methods from the model zoom, we observe several LLMs collapse for the metrics. In contrast, our UrbanLLM significantly outperforms all baseline models on all evaluation metrics. Specifically, UrbanLLM achieved an overall accuracy of 68.3\%, with 95.78\% accuracy in single-task scenarios and 59.08\% in complex real-world problems. By comparison, GPT-4o, the next best-performing model, managed only around 50\% accuracy in spatio-temporal analysis, with other baseline models struggling to effectively complete task decomposition. This demonstrates UrbanLLM's superior capability in handling the intricate demands of urban scenarios.

% However, due to a lack of sufficient training corpus in urban scenarios, LLMs often experience hallucination issues, such as generating tasks that do not exist, during the spatio-temporal analysis stage, which is fundamental in addressing urban-related problems for its role in translating queries into corresponding spatio-temporal sub-tasks to be solved using methods from the model zoo. It can be observed from the results that our UrbanLLM outperforms all the baselines by a large margin on all evaluation metrics. Specifically, UrbanLLM achieved an overall accuracy of 68.3\%, with an accuracy of 95.78\% on single-task analysis and 59.08\% on complex real-world problems. In comparison, GPT-4o, the next best performing model, achieved around 50\% accuracy on spatio-temporal analysis, while other baselines failed to complete the task decomposition.

\begin{table}[!htbp] 
\small
    \centering  
    \caption{Evaluation for Spatio-Temporal Task Analysis}\label{tab:results}
    \vspace{-0.2cm}
    % \begin{adjustbox}% Adjust the table width
        \begin{tabular}{@{}c|c|c|c|c@{}}
            \hline 
            % \toprule
             & Accuracy & Precision & Recall & F1\\
             \hline
             % \midrule
             % \textit{\footnotesize{few-shot}}& & & & \\
             Llama2-7b& 0.18\% & 10.52\% & 8.75\% & 9.18\%\\
             Vicuna-7b-v1.5& 8.44\% & 14.08\% & 13.89\% & 13.95\%\\
             Llama3-8b& 5.31\% & 12.96\% & 15.50\% & 13.08\%\\
             GPT-3.5& 17.95\% & 23.25\% & 22.35\% & 22.54\%\\
             GPT-4o& \underline{49.99\%} & \underline{55.31\%} & \underline{54.42\%} & \underline{54.63\%}\\
             % \midrule
             \hline
             UrbanLLM& \textbf{68.30\%} & \textbf{80.05\%} & \textbf{79.26\%} & \textbf{79.49\%}\\
             % \midrule
             \hline
             \% \textit{Improve}& 36.63\% & 44.73\% & 45.64\% & 45.50\%\\
             \hline
        \end{tabular}
    % \end{adjustbox}
\end{table}

\begin{table}[!htbp] 
\small
    \centering  
    \caption{Evaluation for Spatio-Temporal Single-Task Analysis}\label{tab:simple}
    \vspace{-0.2cm}
    % \begin{adjustbox}% Adjust the table width
        \begin{tabular}{@{}c|c|c|c|c@{}}  
            \hline 
             & Accuracy & Precision & Recall & F1 \\
             \hline
             Llama2-7b & 0.47\% & 0.57\% & 0.57\% & 0.57\% \\
             Vicuna-7b-v1.5& 33.26\% & 33.26\% & 33.26\% & 33.26\%\\
             Llama3-8b& 15.46\% & 17.51\% & 21.19\% & 17.97\%\\
             GPT-3.5 & 13.58\% & 13.70\% & 13.74\% & 13.71\% \\
             GPT-4o & \underline{67.44\%} & \underline{68.56\%} & \underline{68.60\%} & \underline{68.57\%} \\
             \hline 
             UrbanLLM & \textbf{95.78\%} & \textbf{96.78\%} & \textbf{96.84\%} & \textbf{96.80\%} \\
             \hline
             \% \textit{Improve}& 42.02\% & 41.16\% & 41.17\% & 41.17\%\\
             \hline
        \end{tabular}
    % \end{adjustbox}
\end{table}

\begin{table}[!htbp] 
\small
    \centering  
    \caption{Evaluation for Spatio-Temporal Multi-Task Analysis}\label{tab:complex}
    \vspace{-0.2cm}
    % \begin{adjustbox}% Adjust the table width
        \begin{tabular}{@{}c|c|c|c|c@{}}  
            \hline 
             & Accuracy & Precision & Recall & F1\\
             \hline
             Llama2-7b& 0.00\% & 13.80\% & 11.44\% & 12.01\%\\
             Vicuna-7b-v1.5& 0.08\% & 7.62\% & 7.36\% & 7.45\%\\
             Llama3-8b& 1.81\% & 11.36\% & 13.52\% & 11.37\%\\
             GPT-3.5& 19.35\% & 26.40\% & 25.20\% & 25.45\%\\
             GPT-4o& \underline{40.13\%} & \underline{50.89\%} & \underline{49.68\%} & \underline{49.97\%}\\
             \hline 
             UrbanLLM& \textbf{59.08\%} & \textbf{74.47\%} & \textbf{73.40\%} & \textbf{73.71\%}\\
             \hline
             \% \textit{Improve}& 47.22\% & 46.33\% & 47.75\% & 47.51\%\\
             \hline
        \end{tabular}
    % \end{adjustbox}
\end{table}

We also evaluate the performance of UrbanLLM on a human-annotated dataset from 5 spatio-temporal domain experts, as detailed in Table \ref{tab:human}. The results demonstrate that UrbanLLM continues to exhibit substantial improvements over all baseline models. Furthermore, the performance metrics on the human-annotated dataset are comparable to the datasets presented in Table \ref{tab:complex}, which are generated based on the self-instruct method. This consistency in performance suggests that our self-instruct dataset is well-constructed and effectively representative of real-world scenarios, thereby validating the robustness and reliability of UrbanLLM in addressing problems regarding urban activity planning and management.
\begin{table}[!htbp] 
\small
    \centering  
    \caption{Evaluation on Human Annotated Dataset}\label{tab:human}
    \vspace{-0.2cm}
    % \begin{adjustbox}% Adjust the table width
        \begin{tabular}{@{}c|c|c|c|c@{}}  
            \hline 
             & Accuracy & Precision & Recall & F1\\
             \hline
             Llama2-7b& 0.00\% & 14.75\% & 14.00\% & 14.21\%\\
             Vicuna-7b-v1.5& 0.00\% & 5.37\% & 5.08\% & 5.18\%\\
             Llama3-8b& 4.00\% & 13.77\% & 17.67\% & 14.67\%\\
             GPT-3.5& 30.50\% & 37.22\% & 36.88\% & 36.99\%\\
             GPT-4o& \underline{40.50\%} & \underline{49.32\%} & \underline{48.71\%} & \underline{48.89\%}\\
             \hline 
             UrbanLLM& \textbf{55.00\%} & \textbf{74.79\%} & \textbf{74.92\%} & \textbf{74.83\%}\\
             \hline
             \% \textit{Improve}& 35.80\% & 51.64\% & 53.80\% & 53.06\%\\
             \hline
        \end{tabular}
    % \end{adjustbox}
\end{table}
% \vspace{-0.3cm}

\begin{table}[tbp] 
\small
    \centering  
    \caption{Ablation Study of UrbanLLM}\label{tab:ablation}
    % \vspace{-0.2cm}
    % \begin{adjustbox}% Adjust the table width
        \begin{tabular}{@{}c|c|c|c|c@{}}  
            \hline 
             & Accuracy & Precision & Recall & F1\\
             \hline
             UrbanLLM& 68.30\% & 80.05\% & 79.26\% & 79.49\%\\
             \hline 
             w/o SF& 57.02\% & 77.88\% & 76.98\% & 77.24\%\\
             w/o TU& 61.63\% & 78.92\% & 78.07\% & 78.31\%\\
             w/o CU& 62.69\% & 79.10\% & 78.26\% & 78.50\%\\
             \hline
        \end{tabular}
    % \end{adjustbox}
    % \vspace{-0.4cm}
\end{table}

\subsection{Ablation Study}
% \vspace{-0.2cm}
Based on the superiority of UrbanLLM over other established LLMs, we have validated the effectiveness of the training phase. To further demonstrate the contributions of the structured organization of our instructive prompts, we conduct an ablation study on UrbanLLM by removing different components within the spatio-temporal analysis stage. To this end, we define three variants of UrbanLLM  as follows:
% \vspace{-0.2cm}
\begin{itemize}
    % \vspace{-0.5cm}
    \item
    \textbf{w/o SF}: The scenario formulation component is removed from the prompts in the spatio-temporal analysis stage.
    % % % \vspace{-0.2cm}
    \item
    \textbf{w/o TU}: The task understanding component is removed from the prompts in the spatio-temporal analysis stage.
    % % \vspace{-0.2cm}
    \item
    % \vspace{-0.3cm}
    \textbf{w/o CU}: The causal understanding component is removed from the prompts in the spatio-temporal analysis stage.
    % \vspace{-0.2cm}
\end{itemize}
% \vspace{-0.2cm}
The results of different model variants on the four metrics are presented in Table \ref{tab:ablation}. We observe that the removal of each component leads to a decline in performance across all metrics, with the scenario formulation component having the most significant impact. This is likely because the scenario formulation component provides the task scope and the definition of arguments, which are foundations for accurate task decomposition. Moreover, UrbanLLM, which includes all the components in prompts, consistently outperforms all other variants.  This validates the effectiveness of the integration of scenario formulation, task understanding, and causal understanding in inputs to solve urban-related problems. The ablation study confirms that each component is beneficial in model's ability in complex urban activity planning and management.

\subsection{Visualization and Generalization}
\begin{figure}[t]
  \includegraphics[width=\columnwidth]{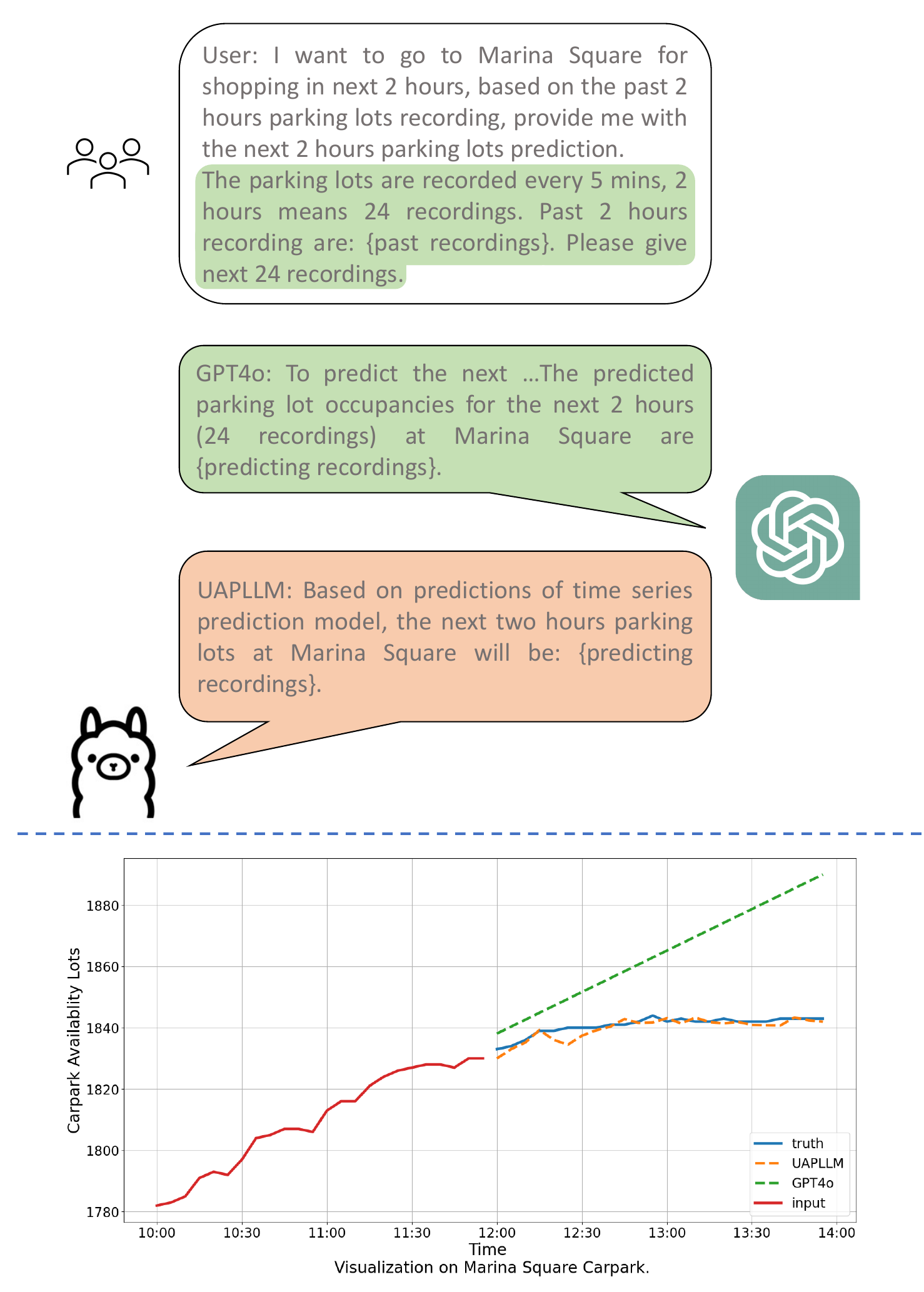}
  \caption{Visualization of responses and results of parking lot occupancy prediction for Marina Square Carpark. GPT-4o needs additional input data highlighted in green for prediction, while UrbanLLM retrieves corresponding data automatically and produces more accurate prediction.}
  % \vspace{-0.5cm}
  \label{fig:predictions}
\end{figure}

% \vspace{-0.2cm}
We further demonstrate the effectiveness of UrbanLLM through a case study focusing on real-world carpark availability prediction problem. In this study, we compare UrbanLLM with the latest GPT-4o model, and their respective responses and predictions are presented in Figure \ref{fig:predictions}. By leveraging a dedicated time series prediction method tailored for the parking domain, UrbanLLM provides robust and accurate predictions. In contrast, GPT-4o, which lacks the capability to to execute specialized spatio-temporal models, produce inaccurate predictions. This comparison shows that UrbanLLM's pipeline yields more reliable outcomes, thereby validating its superior performance in complex urban scenarios.

\begin{figure}[h]
  \centering
  \includegraphics[width=0.7\columnwidth]{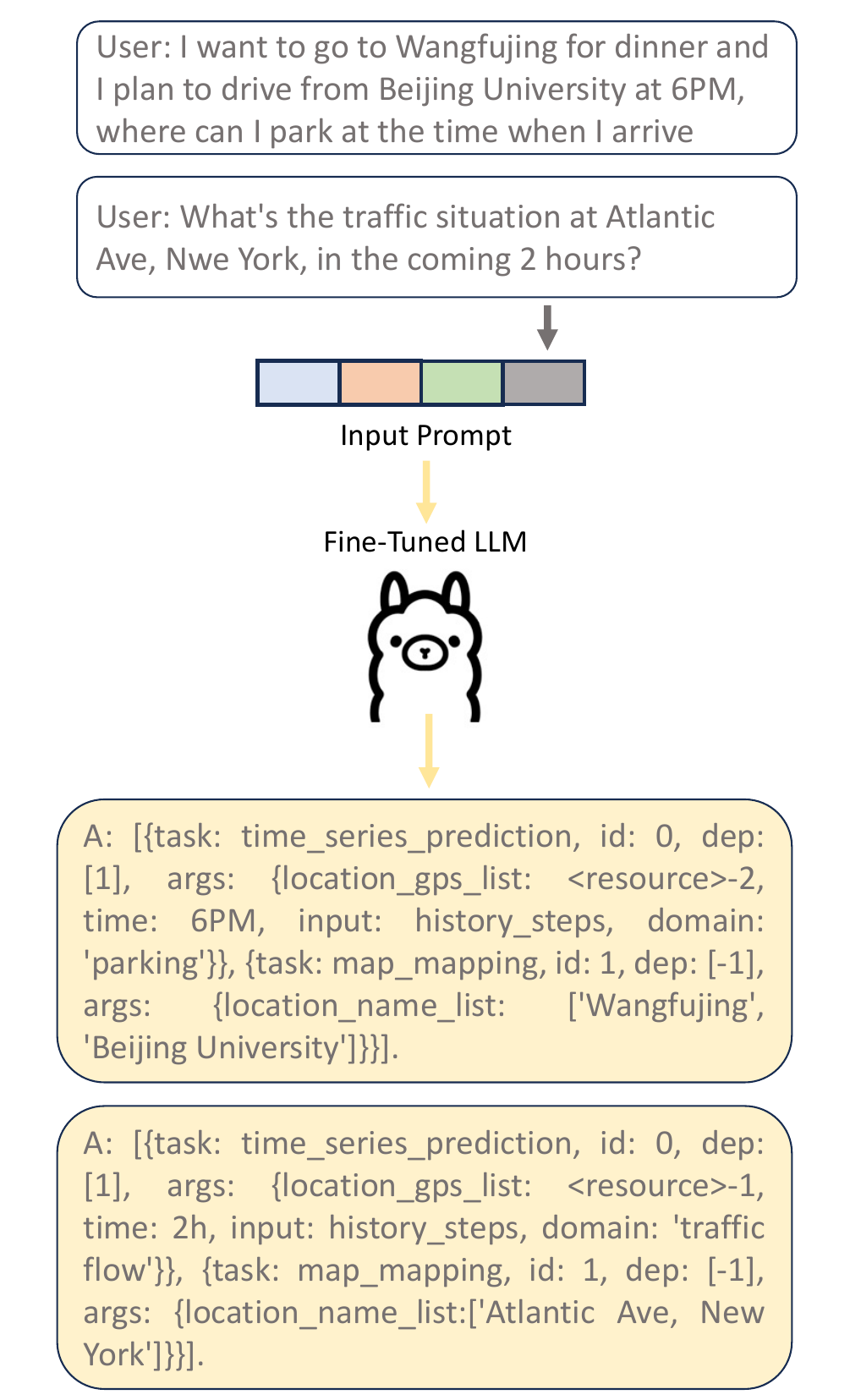}
  \caption{Demonstration of generalization ability across cities in UrbanLLM. The upper part is sample user queries in Beijing and New York City, and the lower part presents the resonable outcomes of spatio-temporal task decomposition.}
  % \vspace{-0.5cm}
  \label{fig:spatial_gene}
\end{figure}

Although initially designed to address urban-related problems specific to Singapore, UrbanLLM are found to exhibit reasonable performance in generalizing to scenarios in other cities, owing to the zero-shot capabilities inherent in LLMs. As illustrated in Figure \ref{fig:spatial_gene}, UrbanLLM effectively decomposes urban-related problems in cities such as Beijing and New York City into relevant spatio-temporal sub-tasks. This generalization ability demonstrates UrbanLLM's versatility and robustness, making it a valuable tool for urban activity planning and management across urban environments.
% % \vspace{-0.4cm}

% % \vspace{-0.4cm}
\section{Conclusions}
% \vspace{-0.3cm}
In this study, we introduce UrbanLLM, a fine-tuned LLM developed to enhance the ability to perform autonomous urban activity planning management. 
After fine-tuning on a corpus of examples of problems in urban scenarios, UrbanLLM learns to decompose new queries into sub-tasks and identifies appropriate spatio-temporal AI models for each sub-task, thereby enhancing the accuracy of urban planning and the efficiency of management processes. Operating through both the learning and the inference phase and three meticulously designed stages, spatio-temporal task analysis, model matching, and results generation, UrbanLLM funcstions as a pipeline to achieve the problem-solving process and produce the response to the given query. Our experimental results demonstrate that UrbanLLM significantly outperforms other LLM models, including Llama-3 and the GPT-4o, in the context of urban activity planning and management tasks by a large margin.

% The complexity inherent in urban planning and management requires innovative solutions for task decomposition and model matching. Traditional spatio-temporal machine learning models fall short in autonomously understanding and managing these intricate tasks, necessitating considerable input from skilled researchers and software engineers. UrbanLLM decomposes complex urban planning and management tasks into subtasks and identifies appropriate spatio-temporal machine learning models for each task, thereby enhancing the accuracy of urban planning and the efficiency of management processes. Operating through both spatio-temporal learning and inference phase and three meticulously designed stages, spatio-temporal task analysis, model matching, and results generation, UrbanLLM ensures smooth automation and integration. Our experimental results validate that UrbanLLM significantly outperforms other LLM-based models, including Llama-3 and the GPT-4o, in the context of urban activity planning and management tasks.

\section*{Limitations}
Despite the promising results demonstrated by UrbanLLM in urban activity planning and management, several limitations need to be acknowledged. \textbf{1) Dependence on Pre-trained Models.} UrbanLLM relies heavily on the performance and capabilities of the underlying pre-trained Llama-2-7B model. While fine-tuning has enhanced its suitability for urban planning tasks, inherent limitations of the base model, such as insufficient geospatial understanding of the target city or specific urban contexts, may still affect outcomes. \textbf{2) Generalization Issues.} The fine-tuning process, while extensive, is based on a specific set of training data and scenarios. This means that UrbanLLM might not generalize well to urban tasks or environments significantly different from those it was trained on. Unexpected urban phenomena or novel planning and management problems may not be adequately addressed by the model. \textbf{3) Resource Intensity.} Although UrbanLLM reduces the need for continuous human intervention, the initial setup and fine-tuning process are resource-intensive. Additionally, as the number of spatio-temporal tasks increases, the self-instruct and fine-tuning process must be repeated, which in turn increases the resource cost. Addressing these limitations in future work will involve enhancing the model’s robustness and geospatial knowledge of the target city, expanding its training datasets to include a more diverse range of scenarios, and developing more efficient fine-tuning techniques. Overcoming these challenges will maximize UrbanLLM’s effectiveness in real-world urban planning applications.

% \section*{Acknowledgments}

\bibliography{custom}

\appendix

\section{Planning and Management Tasks Combinations} \label{sec:taskscombinations}
\lstset{
    basicstyle=\ttfamily\footnotesize,
    breaklines=true,
    columns=fullflexible,
    backgroundcolor=\color{lightgray},
    frame=single,
    numbers=left,
    numberstyle=\tiny\color{gray},
    captionpos=b,
    language=Python
}

We list the 34 planning and management tasks combinations and their corresponding spatio-temporal task decomposition answer in JSON format as follows.

1) Q: I want to go to Jurong East for dinner and will arrive at 7PM, where can I park? Answer:
\begin{lstlisting}
[{task: time_series_prediction, id: 0, dep: [1], args: {location_gps_list: <resource>-1, input: history_steps, domain: 'parking'}}, {task: map_mapping, id: 1, dep: [-1], args: {location_name_list:['Jurong East']}}].
\end{lstlisting}

2) Q: I want to go to Jurong East for dinner and I plan to drive from lake side station at 6PM, where can I park at the time when I arrive? Answer:
\begin{lstlisting}
[{task: time_series_prediction, id: 0, dep: [1], args: {location_gps_list: <resource>-2, time: <resource>-1, input: history_steps, domain: 'parking'}}, {task: arrval_time_estimation, id: 1, dep: [2], args: {location_gps_list: <resource>-2}}, {task: map_mapping, id: 2, dep: [-1], args: {location_name_list:['Jurong East', 'lake side']}}].
\end{lstlisting}

3) Q: Do you have any bicycle parking location recommended nearby Lake Garden? Answer:
\begin{lstlisting}
[{task: recommendation, id: 0, dep: [1], args: {location_gps_list: <resource>-1}, task_specific:['bycycle_parking']}, {task: map_mapping, id: 1, dep: [-1], args: {location_name_list:['Lake Garden']}}].
\end{lstlisting}

4) Q: I would like to go to Starbucks@J-walk, is it here? 3 Gateway Dr. Unit 02-04/04A Westgate, Singapore 608532. Answer:
\begin{lstlisting}
[{task: spatial_relationship_infer, id: 0, dep: [1], args: {location_gps_list: <resource>-1}}, {task: map_mapping, id: 1, dep: [-1], args: {location_name_list:['Starbucks@J-walk', '3 Gateway Dr. #02-04/04A Westgate, Singapore 608532']}}].
\end{lstlisting}

5) Q: I am waiting at the bus stop: 83139. When will be the next No. 15 bus coming? Answer:
\begin{lstlisting}
[{task: bus_arrival, id: 0, dep: [-1], args: {bus_stop: '83139', service_no: 15, task_specific:'next'}}].
\end{lstlisting}

6) Q: I would like to aboard bus no.15 at the bus stop: 83139. How would the bus crowd situation be for the next 30 mins? Answer:
\begin{lstlisting}
[{task: bus_arrival, id: 0, dep: [-1], args: {bus_stop: '83139', service_no: 15, task_specific:'next 30 mins'}}].
\end{lstlisting}

7) Q: My current location is inside Jem shopping centre, where are nearby taxi stands? Answer:
\begin{lstlisting}
[{task: recommendation, id: 0, dep: [1], args: {location_gps_list: <resource>-1},task_specific:['taxi_stand']}, {task: map_mapping, id: 1, dep: [-1], args: {location_name_list:['Jem shopping centre']}}].
\end{lstlisting}

8) Q: My current location is inside Jem shopping centre, how many available taxi arounds my location, like within 2km? Answer:
\begin{lstlisting}
[{task: taxi_availability, id: 0, dep: [1], args: {location_gps_list: <resource>-1},task_specific:['2km']}, {task: map_mapping, id: 1, dep: [-1], args: {location_name_list:['Jem shopping centre']}}].
\end{lstlisting}

9) Q: What's the traffic situation at Serangoon road right now? Answer:
\begin{lstlisting}
[{task: time_series_prediction, id: 0, dep: [1], args: {location_gps_list: <resource>-1}, time: 0, input: history_steps, domain: 'traffic speed'}, {task: map_mapping, id: 1, dep: [-1], args: {location_name_list:['Serangoon road']}}].
\end{lstlisting}

10) Q: What's the traffic situation at Serangoon road in the coming 2 hours? Answer:
\begin{lstlisting}
[{task: time_series_prediction, id: 0, dep: [1], args: {location_gps_list: <resource>-1}, time: 2h, input: history_steps, domain: 'traffic speed'}, {task: map_mapping, id: 1, dep: [-1], args: {location_name_list:['Serangoon road']}}].
\end{lstlisting}

11) Q: What's the traffic situation at PIE express way in the coming week? Answer:
\begin{lstlisting}
[{task: long_time_series_prediction, id: 0, dep: [1], args: {location_gps_list: <resource>-1}, time: 1w, input: history_steps, domain: 'traffic speed'}, {task: map_mapping, id: 1, dep: [-1], args: {location_name_list:['PIE express way']}}].
\end{lstlisting}

12) Q: As a land and traffic regulator, can you tell whether there are any abnormal traffic speed with Jurong Area? Answer:
\begin{lstlisting}
[{task: time_series_anomaly_detection, id: 0, dep: [1], args: {location_gps_list: <resource>-1}, input: history_steps, domain: 'traffic speed'}, {task: map_mapping, id: 1, dep: [-1], args: {location_name_list:['Jurong Area']}}].
\end{lstlisting}

13) Q: As a land and traffic regulator, can you tell whether there are any abnormal traffic speed in whole Singapore right now? Answer:
\begin{lstlisting}
[{task: time_series_anomaly_detection, id: 0, dep: [-1], args: {input: history_steps, domain: 'traffic speed'}}].
\end{lstlisting}

14) Q: As a land and traffic regulator, can you infer the missing traffic speed values with Jurong Area? Answer:
\begin{lstlisting}
[{task: time_series_imputation, id: 0, dep: [1], args: {location_gps_list: <resource>-1}, input: history_steps, domain: 'traffic speed'}, {task: map_mapping, id: 1, dep: [-1], args: {location_name_list:['Jurong Area']}}].
\end{lstlisting}

15) Q: As a land and traffic regulator, can you infer the missing traffic speed values in whole Singapore right now? Answer:
\begin{lstlisting}
[{task: time_series_imputation, id: 0, dep: [-1], args: {input: history_steps, domain: 'traffic speed'}}].
\end{lstlisting}

16) Q: What is the weather nearby NTU right now, is it going to rain? Answer:
\begin{lstlisting}
[{task: time_series_prediction, id: 0, dep: [1], args: {location_gps_list: <resource>-1}, time: 0, input: history_steps, domain: 'precipitation'}, {task: map_mapping, id: 1, dep: [-1], args: {location_name_list:['NTU']}}].
\end{lstlisting}

17) Q: What is the weather nearby NTU right now, is it going to rain for the next 2 hours? Answer:
\begin{lstlisting}
[{task: time_series_prediction, id: 0, dep: [1], args: {location_gps_list: <resource>-1}, time: 2h, input: history_steps, domain: 'precipitation'}, {task: map_mapping, id: 1, dep: [-1], args: {location_name_list:['NTU']}}].
\end{lstlisting}

18) Q: What is the air quality nearby NTU right now? Answer:
\begin{lstlisting}
[{task: time_series_prediction, id: 0, dep: [1], args: {location_gps_list: <resource>-1}, time: 0, input: history_steps, domain: 'air'}, {task: map_mapping, id: 1, dep: [-1], args: {location_name_list:['NTU']}}].
\end{lstlisting}

19) Q: What is the weather nearby NTU for the next 2 hours? Answer:
\begin{lstlisting}
[{task: time_series_prediction, id: 0, dep: [1], args: {location_gps_list: <resource>-1}, time: 2h, input: history_steps, domain: 'air'}, {task: map_mapping, id: 1, dep: [-1], args: {location_name_list:['NTU']}}].
\end{lstlisting}

20) Q: As a land and traffic regulator, can you infer the missing parking records for HDB carpark 655? Answer:
\begin{lstlisting}
[{task: time_series_imputation, id: 0, dep: [1], args: {location_gps_list: <resource>-1}, input: history_steps, domain: 'parking'}, {task: map_mapping, id: 1, dep: [-1], args: {location_name_list:['HDB carpark 655']}}].
\end{lstlisting}

21) Q: As a land and traffic regulator, can you infer the missing parking records all singapore residential carparks? Answer:
\begin{lstlisting}
[{task: time_series_imputation, id: 0, dep: [1], args: {input: history_steps, domain: 'parking', task_specific: ['residential carparks']}}].
\end{lstlisting}

22) Q: As a land and traffic regulator, can you tell whether there are any abnormal parking records for HDB carpark 655? Answer:
\begin{lstlisting}
[{task: time_series_anomaly_detection, id: 0, dep: [1], args: {location_gps_list: <resource>-1}, input: history_steps, domain: 'parking'}, {task: map_mapping, id: 1, dep: [-1], args: {location_name_list:['HDB carpark 655']}}].
\end{lstlisting}

23) Q: As a land and traffic regulator, can you tell whether there are any abnormal parking records form all singapore residential carparks? Answer:
\begin{lstlisting}
[{task: time_series_anomaly_detection, id: 0, dep: [1], args: {input: history_steps, domain: 'parking', task_specific: ['residential carparks']}}].
\end{lstlisting}

24) Q: As a land and traffic regulator, can you infer traffic risk for the next one week? Answer:
\begin{lstlisting}
[{task: event_prediction, id: 0, dep: [1], args: {input: history_steps, time: 1w, domain: 'traffic accident'}}].
\end{lstlisting}

25) Q: As a land and traffic regulator, can you infer traffic risk within Jurong area for the next one week? Answer:
\begin{lstlisting}
[{task: event_prediction, id: 0, dep: [1], args: {location_gps_list: <resource>-1}, input: history_steps, time: 1w, domain: 'traffic accident'}, {task: map_mapping, id: 1, dep: [-1], args: {location_name_list:['Jurong area']}}].
\end{lstlisting}

26) Q: As a data network provider, you require abundant user trajectory data to improve signal service. Please infer the missing trajectory for the provided trajectory data? Answer:
\begin{lstlisting}
[{task: trajectory_completion, id: 0, dep: [-1], args: {input: trajectory_records, domain: 'user trajectory'}}].
\end{lstlisting}

27) Q: As a data network provider, you require abundant user trajectory data to improve signal service. Based on the trajectory provided, predict the next day user trajectory? Answer:
\begin{lstlisting}
[{task: trajectory_prediction, id: 0, dep: [-1], args: {input: trajectory_records, time: 1d, domain: 'user trajectory'}}].
\end{lstlisting}

28) Q: I would like to have dinner with my girlfriend at Jurong East at 7PM, can you recommend a Japanese restaurant with parking space? Answer:
\begin{lstlisting}
[{task: time_series_prediction, id: 0, dep: [2], args: {location_gps_list: <resource>-2, time: 7PM, input: history_steps, domain: 'parking'}}, {task: recommendation, id: 1, dep: [2], args: {location_gps_list: <resource>-2, task_specific: 'Japanese restaurant'}}, {task: map_mapping, id: 2, dep: [-1], args: {location_name_list:['Jurong East']}}].
\end{lstlisting}

29) Q: I would like to have dinner with my girlfriend at Jurong East, we will depart from lake side at 5PM, can you recommend a Japanese restaurant with available parking space when we arrive? Answer:
\begin{lstlisting}
[{task: time_series_prediction, id: 0, dep: [2], args: {location_gps_list: <resource>-2, time: 7PM, input: history_steps, domain: 'parking'}}, {task: recommendation, id: 1, dep: [2], args: {location_gps_list: <resource>-2, task_specific: 'Japanese restaurant'}}, {task: map_mapping, id: 2, dep: [-1], args: {location_name_list:['Jurong East']}}].
\end{lstlisting}

30) Q: I would like to play basketball at NTU SRC outdoor courts with my friends. I will drive to NTU and arrive around 7PM, do you have any suggestions? Answer:
\begin{lstlisting}
[{task: time_series_prediction, id: 0, dep: [4], args: {location_gps_list: <resource>-4, time: 7PM, input: history_steps, domain: 'parking'}}, {task: time_series_prediction, id: 1, dep: [4], args: {location_gps_list: <resource>-4, time: 7PM, input: history_steps, domain: 'air'}}, {task: time_series_prediction, id: 2, dep: [4], args: {location_gps_list: <resource>-4, time: 7PM, input: history_steps, domain: 'precipitation'}}, {task: recommendation, id: 3, dep: [4], args: {location_gps_list: <resource>-4, task_specific: 'Japanese restaurant'}}, {task: map_mapping, id: 4, dep: [-1], args: {location_name_list:['NTU SRC outdoor courts']}}].
\end{lstlisting}

31) Q: I would like to play basketball at NTU SRC outdoor courts with my friends. I will drive from lake side MRT at around 7PM to NTU, do you have any suggestions including weather and parking space? Answer:
\begin{lstlisting}
[{task: time_series_prediction, id: 0, dep: [5], args: {location_gps_list: <resource>-5, time: <resource>-4, input: history_steps, domain: 'parking'}}, {task: time_series_prediction, id: 1, dep: [5], args: {location_gps_list: <resource>-5, time: <resource>-4, input: history_steps, domain: 'air'}}, {task: time_series_prediction, id: 2, dep: [5], args: {location_gps_list: <resource>-5, time: <resource>-4, input: history_steps, domain: 'precipitation'}}, {task: recommendation, id: 3, dep: [5], args: {location_gps_list: <resource>-5, task_specific: 'Japanese restaurant'}}, {task: arrval_time_estimation, id: 4, dep: [5], args: {location_gps_list: <resource>-5}}, {task: map_mapping, id: 5, dep: [-1], args: {location_name_list:['NTU SRC outdoor courts', 'lake side MRT']}}].
\end{lstlisting}

32) Q: I would like find a gym for exercise and having dinner with my friends, we prefer the western food nearby the gym. Can you help to plan out the activities? Answer:
\begin{lstlisting}
[{task: recommendation, id:0, dep: [1], args: {location_gps_list: <resource>-1, task_specific: 'western food'}}, {task: recommendation, id: 1, dep: [-1], args: {task_specific: 'gym'}}].
\end{lstlisting}

33) Q: I would like find a gym for exercise and having dinner with my friends, we prefer the western food nearby the gym. Can you help to plan out the activities within Jurong area? Answer:
\begin{lstlisting}
[{task: recommendation, id:0, dep: [1], args: {location_gps_list: <resource>-1, task_specific: 'western food'}}, {task: recommendation, id: 1, dep: [2], args: {location_gps_list: <resource>-2, task_specific: 'gym'}}, {task: map_mapping, id: 2, dep: [-1], args: {location_name_list:['Jurong area']}}].
\end{lstlisting}

34) Q: I would like to find a gym for exercise and having dinner with my friends, we prefer the western food nearby the gym. Can you help to plan out the activities within Jurong area? I will arrive at 5PM to 5:30PM, please let me know where to park my car as well. Answer:
\begin{lstlisting}
[{task: time_series_prediction, id: 0, dep: [1], args: {location_gps_list: <resource>-1, time: 2h, input: history_steps, domain: 'parking'}}, {task: recommendation, id:1, dep: [2], args: {location_gps_list: <resource>-2, task_specific: 'western food'}}, {task: recommendation, id: 2, dep: [3], args: {location_gps_list: <resource>-3, task_specific: 'gym'}}, {task: map_mapping, id: 3, dep: [-1], args: {location_name_list:['Jurong area']}}].
\end{lstlisting}

\section{Task Understanding Prompt} \label{sec:taskunderstanding}
In this section, we present a full task understanding prompt used instruction-tuning of UrbanLLM.

To better understand each spatial-temporal task, here is the explanation and numbering, along with the corresponding examples: 

1) Long Time Series Prediction: This task involves forecasting future values in a time series over a long horizon. It is typically used for long-term planning and trend analysis in various domains, such as weather forecasting, economic forecasting, and demand planning. 

2) Time Series Prediction: This task focuses on predicting future values in a time series over a shorter horizon compared to long time series prediction. It is commonly used for short-term forecasts like daily stock prices, temperature forecasts, or short-term sales predictions. 

3) Event Prediction: This task involves predicting the occurrence of specific events based on historical data. Examples include predicting natural disasters, equipment failures, or social events like concerts or sports games. 

4)Trajectory Completion: This task involves completing missing parts of a trajectory based on observed segments. It is useful in applications like tracking moving objects, filling in missing GPS data, or reconstructing incomplete travel routes. 

5)Trajectory Prediction: This task involves forecasting the future path of a moving object based on its past trajectory. Applications include predicting the movement of vehicles, pedestrians, or animals. 

6)Time Series Anomaly Detection: This task involves identifying unusual patterns or outliers in time series data that deviate from expected behavior. It is used in applications like fraud detection, fault detection in machinery, and monitoring traffic conditions. 

7) Time Series Imputation: This task involves filling in missing values in time series data to ensure completeness and consistency. It is crucial for maintaining data quality in various applications like traffic records and climate data.

8)Arrival Time Estimation: This task involves predicting the arrival time of a vehicle or person at a specific location based on current and historical data. It is commonly used in transportation systems for buses, trains, and delivery services. 

9) Taxi Availability Prediction: This task involves predicting the availability of taxis in specific areas at given times. It helps optimize taxi dispatching and improve service for passengers by anticipating demand and ensuring timely availability.  

10) Map Mapping: This task involves mapping addresses to GPS locations and mapping GPS locationsback to addresses .

11) Bus Arrival: This task involves predicting the arrival times of buses at specific stops based on real-time data and historical patterns. It enhances the efficiency of public transportation systems by providing accurate and timely information to commuters. 

12) Spatial Relationship Inference: This task involves deducing spatial relationships between different entities or locations. It is used in urban planning to understand spatial dependencies and interactions, such as proximity analysis, clustering, and spatial correlations. 

13) Recommendation: This task involves suggesting items or actions to users based on their preferences and historical behavior. Applications include recommending points of interest, routes, or services in urban planning.

\section{Spatio-Temporal Model Description} \label{sec:modelmatching}
We demonstrate candidate model description for time series predictions as below: 

[{'model id':1, 'model name':'DCRNN', 'data domain':'road traffic speed', 'description':'DCRNN is a deep learning architecture designed for traffic forecasting tasks, particularly in urban areas. It combines techniques from convolutional and recurrent neural networks to effectively capture spatial and temporal dependencies in traffic data. The model takes advantage of the graph structure of traffic data, where nodes represent different locations (such as intersections or sensors) and edges represent connections between these locations.'}, {'model id':2, 'model name':'SAGDFN', 'data domain':'road traffic speed, carpark availability lots', 'description':'SAGDFN: A Scalable Adaptive Graph Diffusion Forecasting Network for Multivariate Time Series Forecasting aims to provide accurate multivariaate time series predictions for both normal and larger datasets and datasets span various times series domains such as traffic speed to carpark availability lots.'}, {'model id':3, 'model name':'AGCRN', 'data domain':'road traffic speed', 'description':'Adaptive graph convolutional recurrent network for traffic forecasting takes advantage of the graph structure of traffic data to assist in traffic speed predictions.'}]

\section{Evaluation Metrics} \label{sec:evluationmetrics}

\noindent \textbf{Precision.}
Macro precision is the average precision for each sample.

For each task \(i\):
\[
\text{Precision}_i = \frac{TP_i}{TP_i + FP_i}
\]

Where:
- \(TP_i\) is the number of true positives for task \(i\)
- \(FP_i\) is the number of false positives for task \(i\)

Macro precision:
\[
\text{Macro Precision} = \frac{1}{N} \sum_{j=1}^{N} (\frac{1}{C} \sum_{i=1}^{C} \text{Precision}_i)_j
\]

\noindent \textbf{Recall.}
Macro recall is the average recall for each sample.

For each task \(i\):
\[
\text{Recall}_i = \frac{TP_i}{TP_i + FN_i}
\]

Where:
- \(TP_i\) is the number of true positives for task \(i\)
- \(FN_i\) is the number of false negatives for task \(i\)

Macro recall:
\[
\text{Macro Recall} = \frac{1}{N} \sum_{j=1}^{N} (\frac{1}{C} \sum_{i=1}^{C} \text{Recall}_i)_j
\]

\noindent \textbf{F1 Score.}
Macro F1 score is the average F1 score for each sample.

For each class \(i\):
\[
\text{F1}_i = 2 \cdot \frac{\text{Precision}_i \cdot \text{Recall}_i}{\text{Precision}_i + \text{Recall}_i}
\]

Macro F1 score:
\[
\text{Macro F1 Score} = \frac{1}{N} \sum_{j=1}^{N} (\frac{1}{C} \sum_{i=1}^{C} \text{F1}_i)_j
\]

Where \(N\) is the total number of samples and \(N\) is the total number tasks types.

\end{document}